\documentclass[runningheads]{llncs}

\usepackage[colorlinks=true,linkcolor = blue,urlcolor = blue]{hyperref}
\usepackage{amsmath}
\usepackage{amssymb}
\usepackage{graphicx}
\usepackage{siunitx}
\usepackage[accsupp]{axessibility}
\usepackage[table]{xcolor}

\begin{document}
\pagestyle{headings}
\mainmatter
\def\ECCVSubNumber{6195}

\title{Spline-NeRF: $C^0$-Continuous Dynamic NeRF}
\author{Julian Knodt}
\institute{Princeton University}

\maketitle

\begin{abstract}
The problem of reconstructing continuous functions over time is important for problems such as reconstructing moving scenes, and interpolating between time steps.
Previous approaches that use deep-learning rely on regularization to ensure that reconstructions are approximately continuous, which works well on short sequences. As sequence length grows, though, it becomes more difficult to regularize, and it becomes less feasible to learn only through regularization.

We propose a new architecture for function reconstruction based on classical Bezier splines, which ensures $C^0$ and $C^1$-continuity, where $C^0$ continuity is that $\forall c:\lim\limits_{x\to c} f(x) = f(c)$, or more intuitively that there are no breaks at any point in the function. In order to demonstrate our architecture, we reconstruct dynamic scenes using Neural Radiance Fields, but hope it is clear that our approach is general and can be applied to a variety of problems. We recover a Bezier spline $B(\beta, t\in[0,1])$,
parametrized by the control points $\beta$. Using Bezier splines ensures reconstructions have $C^0$ and $C^1$ continuity, allowing for guaranteed interpolation over time. We reconstruct $\beta$ with a multi-layer perceptron (MLP), blending machine learning with classical animation techniques. All code is available at
\url{https://github.com/JulianKnodt/nerf_atlas}, and datasets are from prior work.

\end{abstract}

\begin{figure}[!ht]
    \centering
    \includegraphics[width=\textwidth]{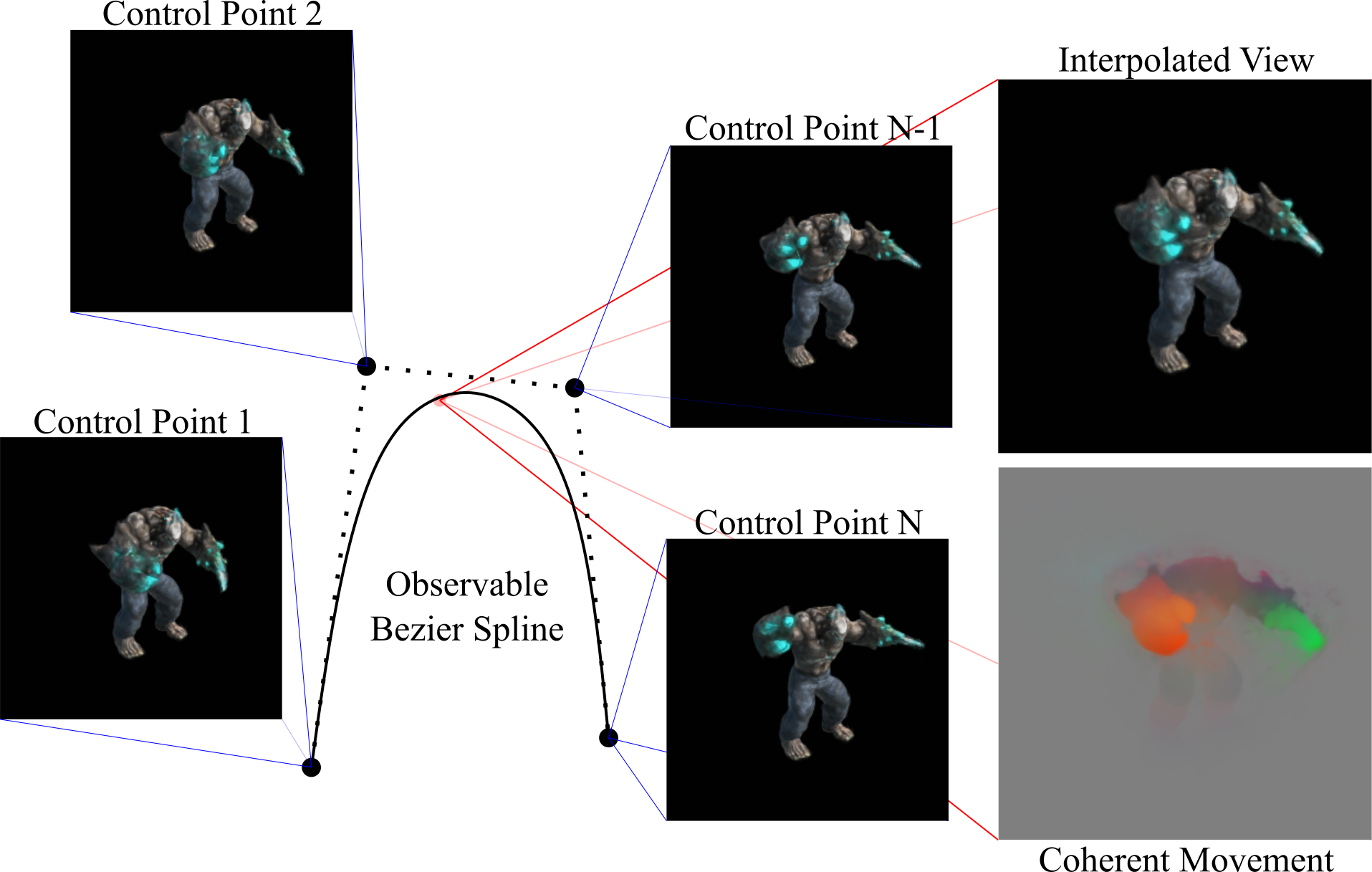}
    \caption{
    \label{fig:intro_figure}
    \textbf{Overview of approach.} Conceptually, our approach models movement as interpolation along a high-dimensional bezier spline, where each control point is an object. This is done by learning a function which maps from $x\in\mathbb{R}^3$ to a Bezier spline with $N$ control points. In the figure above, only the first and last control points are visible while rendering. Using this classical formulation, we impose a strong prior on continuity of movement, while enforcing that it is smooth. We compare our model to an implementation of NR-NeRF~\cite{tretschk2021nonrigid}, and find our model is on par qualitatively and produces more coherent movement. A qualitative difference can also be seen in videos comparing the two reconstructions, due to the difference in the velocity of movement.
    }
    \vspace{-3mm}
\end{figure}

\section*{Introduction}

Learning continuous, smooth, functions is a key problem in machine learning, as many problems
are framed as finding good interpolations between a few data points, where we define continuity as $\lim\limits_{x\to c}f(x) = f(c)$.
The continuity of a learned model is not guaranteed, and is enforced through regularization of the output, either by having enough training samples in a short time, or
using regularization such as total variation across time. Since consistency is dependent on data and regularization and cannot be guaranteed, it is empirically demonstrated. This leads to difficulties in interpolating between sparse training data, and the possibility for sudden changes in output between training points, such as suddenly jumping from one frame to the next in a learned video.
To enforce continuity, we are interested in recovering functions which guarantee $C^0$ continuity. $C^0$ continuity is a useful property for many tasks, such as in reconstructing movement to ensure that an object cannot warp between two points instantaneously. We are also interested in $C^1$ continuity, or that the derivative of a function is continuous on some domain. This is because it is not physically possible for an object to instantly change its velocity, therefore reconstructions must have $C^1$ continuity for plausible movement.

To demonstrate how to construct functions with these properties, we tackle the problem of dynamic scene reconstruction using NeRF~\cite{mildenhall2020nerf}, which is a recent method for reconstructing scenes. Following previous work, we define a static scene, which is referred to as the canonical scene, and a model which can produce deformations to the canonical scene. Our approach is a small modification to prior work: to represent the canonical scene, we use a NeRF~\cite{mildenhall2020nerf}, and the deformation model used to model movement is our proposed learned approach. The use of deformation networks has been shown to be effective at reconstructing synthetic scenes with movement as in D-NeRF~\cite{pumarola2020dnerf} and real scenes in NR(non-rigid)-NeRF~\cite{tretschk2021nonrigid}. These works show convincing
reconstructions of moving scenes, allowing for novel view synthesis from video. These methods do not have an analytic form, relying purely on learned components, but we would like to be able to analyze and modify the movement. For example, an application may want to cluster movement, or change directions, but prior work does not immediately provide a method for doing so. In contrast, classical animation tools are designed to allow for control of movement, but their use has not been explored in prior work.

To this end, we look to existing tools in animation for creating realistic movement while allowing a high-degree of control for artists and animators. For example, there are tools such as keyframing and splines, which allow animators to construct movement with a small set of tunable knobs. Despite few degrees of freedom, these tools allow artists and animators to breathe life into animation with a high degree of control. In addition, mathematical constructs such as splines have also been thoroughly studied to understand their behaviour and how they can be manipulated, and thus are readily modifiable in post-processing.

Thus, we use the animation techniques of Bezier splines as a method to enforce continuous interpolation in Dynamic NeRF, finding that we are able to get comparable performance without additional cost in memory, and the desired properties of $C^0, C^1$ continuity.
In summary, our contributions are as follows:

\begin{enumerate}
    \item A general architecture for $C^0, C^1$ continuity over a continuous domain.
    \item An application of this architecture, building on NR-NeRF to enforce continuity of movement with an analytic form with negligible computational cost, while performing on par with the original.
\end{enumerate}

\begin{figure}
    \vspace{-6mm}
    \centering
    \begin{minipage}[c]{0.5\textwidth}
    \includegraphics[width=\textwidth]{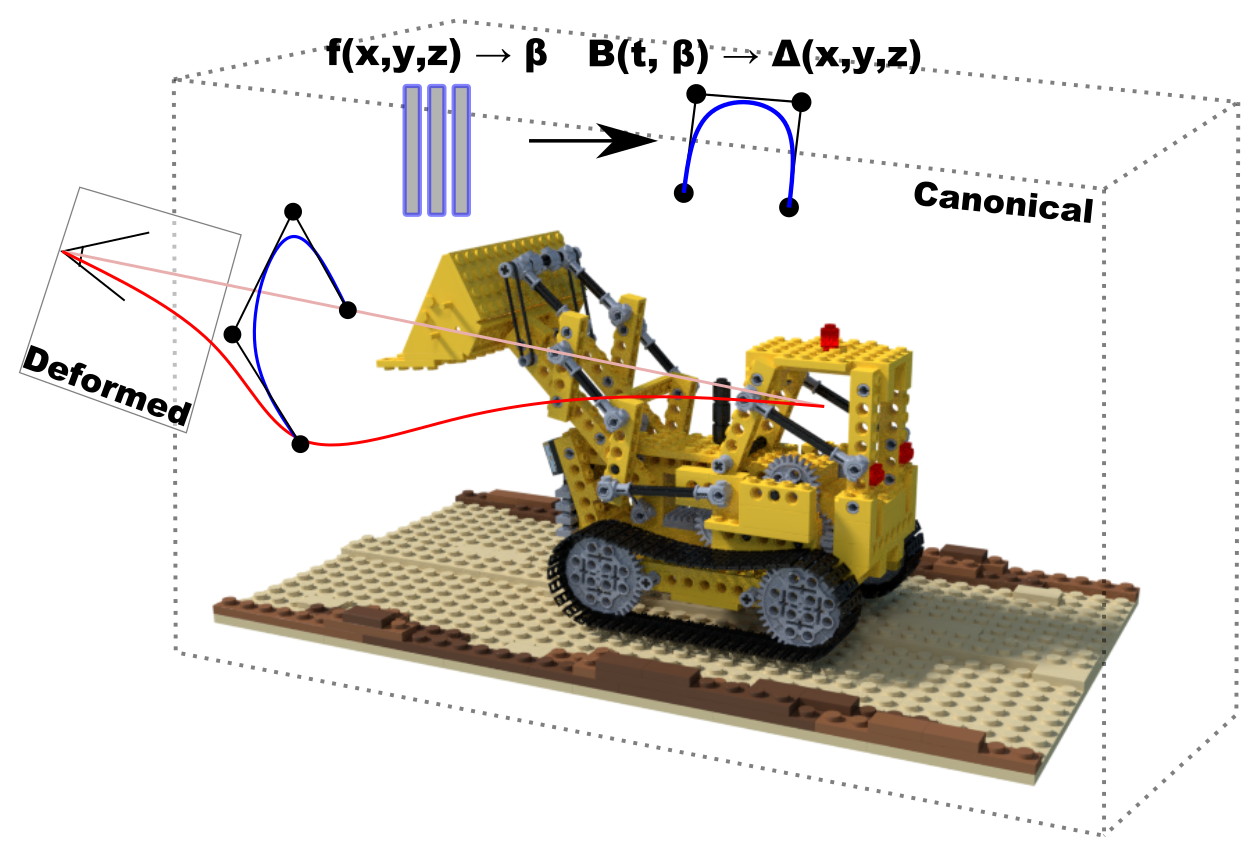}
    \end{minipage}
    \begin{minipage}[c]{0.45\textwidth}
    \caption{
        \label{fig:arch_diagram}
        \textbf{Spline-based Dynamic NeRF}. Instead of using an MLP to directly predict ray-bending, $x' = \Delta(\text{MLP}(x, t)) + x$,  at a given time, we predict a set of Bezier spline control points. Then, we use the Bezier spline defined by these points to interpolate position based on time: \newline $x' = \Delta(B(\beta,t)) + x$, where $B$ is a Bezier spline parametrized by $\beta = \text{MLP(x)}$.
    }
    \end{minipage}
    \vspace{-10mm}
\end{figure}

\section*{Related Work \& Background}

\subsection*{Static Scene Reconstruction}

Before considering scenes with movement or other changing parameters, it's useful to consider static scenes with no transformations.
Static scene reconstruction is the problem of reconstructing a 3D scene from a set of 2D views of a non-changing scene, and Neural Radiance Fields (NeRFs)~\cite{mildenhall2020nerf} are a recent technique achieving this. NeRFs model a scene as a continuous volume of varying density, which has view-dependent color, allowing for reconstruction of highly-detailed scenes. NeRF is based on traditional volume rendering techniques:
\begin{align}\label{eq:nerf}
I(r) &= \int_{t_n}^{t_f} T(t, r) \sigma(r(t)) c(r(t), r_d)dt \nonumber \\
T(t, r) &= \exp(-\int_{t_n}^{t} \sigma(r(s))ds)
\end{align}
\noindent
where $I(r)$ is the illumination along camera ray $r(t) = r_o + r_d t$, $r_o, r_d$ are the ray origin and direction respectively, and $t$ is some positive distance along the ray. NeRFs are able to accurately reconstruct high-frequency features by recovering $\sigma$ and $c$, the density and view-dependent color at a given point by modelling them as multi-layer perceptrons (MLPs) with a positional encoding scheme that can differentiate between nearby points in space. NeRFs evaluate the above equations by performing ray-marching and computing $T(j,r) = \sum -\exp\sigma_i c_i$, by partitioning the ray into evenly spaced bins and sampling randomly from within each bin. There has been a plethora of work exploring NeRF and extensions which permit capturing more variance.

These extensions to NeRF include optimizations on the encoding for differentiating positions in space~\cite{tancik2020fourfeat}, better sampling approaches~\cite{barron2021mipnerf}, faster training~\cite{yu2021plenoxels}, and more~\cite{sitzmann2019siren,bi2020neural,srinivasan2020nerv,boss2021nerd}. The underlying canonical model is crucial to our formulation of dynamic NeRF, and for this we use SIREN~\cite{sitzmann2019siren} without a coarse-to-fine approach to get high-frequency details.

\subsection*{Dynamic NeRF Reconstruction}

Dynamic scene reconstruction builds on static scene reconstruction, removing the assumption that all views are under the same condition, such as having the same lighting or that nothing has moved, since NeRFs were designed to only handle static scenes, and thus cannot handle changes between frames.
In order to model dynamic scenes, there have been two diverging approaches.

One kind of approach directly models the transformation in the time domain, by learning a function $\sigma(x,t)=f(x\in\mathbb{R}^3, t\in[0,1])$, which include works such as HyperNeRF~\cite{park2021hypernerf}, NeRFies~\cite{park2021nerfies}, Space-Time Invariant Irradiance Fields~\cite{xian2021space}, and others~\cite{Wang_2021_CVPR,du2021nerflow}. By directly modelling the variation of the density, these methods are able to reconstruct large deformations in latent spaces and reconstruct a wide variety of transformations from a single radiance field. These often allow for new transformations in some learned space between similar views, allowing for warping and interpolation between observed views.

The other kind of approach models movement directly as translation, preventing changes in density or view-dependent effects. NeRFs are not able to move the objects inside the scene since we can only evaluate the NeRF at a given $x\in\mathbb{R}^3$. Instead we bend the rays, warping what is visible from a given view. This is essentially a perspective shift of a transformation of the space being rendered: instead of moving an object that should be seen by ray $r$, we warp ray $r$ such that it sees the object. The equation for density is defined as $\sigma(x,t)=f(x+\Delta(x,t))$. This formulation enforces a coherent canonical representation, while directly modelling movement, and has been shown to be able to reconstruct both synthetic scenes with D-NeRF~\cite{pumarola2020dnerf} and real scenes in NR-NeRF~\cite{tretschk2021nonrigid}. There has also been work on recovering movements of multiple NeRFs whose bounding boxes move within a scene, such as in \cite{dynamicSceneGraphs}, but our work diverges from that approach as we are interested in reconstructing movement within one instance of a NeRF.

The benefits of directly including time as a function in the NeRF are that we are able to represent a broader class of functions: every time step or point in latent space may be fully distinct from others. On the other hand, explicit warping lends itself to smoothness between frames and accurate reconstruction of the physical process. Our approach falls into the warping category, as we are interested in accurately reconstructing smooth movement as opposed to generalizing over many classes of transformations.

As an aside, we note that a lot of prior work related to dynamic NeRFs have not been peer-reviewed~\cite{pumarola2020dnerf,li2021neural,park2021nerfies,neural3dViewSynthesis}, but these works still have a significant impact, especially since the field moves quickly.

\subsection*{Bezier Curves}
\begin{figure}
    \centering
    \includegraphics[width=0.2\textwidth]{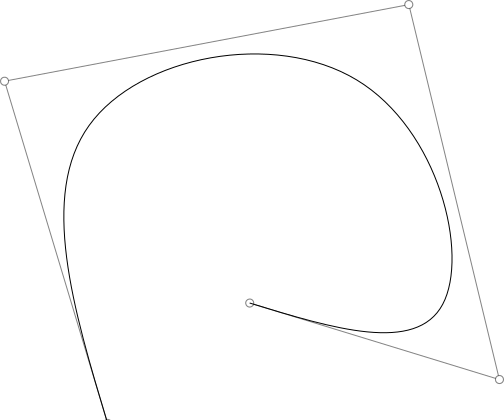}
    \caption{
        \textbf{Example of a Bezier Spline}.
        Bezier splines are a low-dimensional polynomial for smooth interpolation between a few control points. Our method learns control points to produce smooth movement and induce a prior on continuity, drawing inspiration from animation and drawing software.
    }
    \label{fig:bezier_diagram}
\end{figure}

Bezier curves refer to a specific set of polynomials parametrized by a set of control points. They are most commonly represented as cubics: $f(t) = a(1-t)^3 + 3b(1-t)^2t + 3c(1-t)t^2 + dt^3$,
where $t\in[0,1]$ is the variable we are interested in interpolating over, and $a,b,c,d$ are ``control points'' of the function. An example of a Bezier spline (we will use spline and curve interchangeably from here on) is shown in Figure~\ref{fig:bezier_diagram}. The general formulation for
the Bezier basis functions is defined as $B^n(t) = \sum\limits^n_{i=0} {n \choose i} (1-t)^{n-i} t^i$ where $n$
is the degree of the Bezier polynomial. In order to control the Bezier curve, we use control points $P_i$, which control the shape of the curve: $B^n(t) = \sum\limits_{i=0}^n P_i {n \choose i} (1-t)^{n-i} t^i$, where $P_i\in\mathbb{R}^3$ for 3D movement. For a more comprehensive guide on Bezier splines, we refer the reader to a more
\href{https://pomax.github.io/bezierinfo/index.html}{complete reference}~\cite{bezier_primer}\footnote{While this is a not a published, peer-reviewed source, the author found it to be the most well-written, free, and comprehensive resource available.}.

\section*{Method}

Our approach imposes structure on top of machine learning
approaches, enforcing properties on the reconstructed values. For functions \[ f(p, t)\to\mathbb{R}^n, t\in[0,1], \] we decompose $f(p,t)$ into a function: 
\begin{align}
    f(p)\to\mathbb{R}^{n\times O}, B_O(f(p), t)\to\mathbb{R}^n
\end{align}
Where
$O$ is the order of the Bezier spline, $f(p)$ is the learned control points, and $B_O(f(p), \cdot)$ is the evaluation
of the $O$th order Bezier spline with control points defined by $f(p)$.

\subsection*{Architecture}

For dynamic NeRF, we define $f(p)$ as
$\text{MLP}(x,y,z)\to\mathbb{R}^{3\times O}$. We ray march from a camera with known position and
view direction through the scene, and at every point we compute the set of control points for a Bezier curve. We then evaluate the Bezier curve at a given time, and deform the ray by the result, producing some $\Delta_0(x)$. The number of spline points for the Bezier curve is a  hyperparameter, and our experiments use 5 spline points. In order to evaluate the Bezier curve in a numerically stable way, we use De Casteljau's algorithm. 

De Casteljau's algorithm evaluated at time $t$ is defined by the recurrence relation:
\begin{align}
  \beta_i^{(0)} &= \beta_i \nonumber \\
  \beta_i^{j} &= (1-t)\beta_i^{(j-1)} + t\beta_{i-1}^{(j-1)}
\end{align}
which can be thought of as linearly interpolating between adjacent control points until there is only a single fixed point. This takes $O(n^2)$ operations to evaluate, where $n$ is the number of control points. For a small $n$, i.e. 5 spline points which is what we evaluate on, this is negligible.

We are also interested in constructing a reasonable canonical NeRF, and without loss of generality select $t = 0$ to be canonical. From this, we are interested in Bezier Curves where $B_O(0) = \overrightarrow{0}$. This can be achieved in two different ways, either by assigning $p_0 = \overrightarrow{0}$, and only computing the other control points: $f(p) = p_{1\cdots O-1}$. Then, we can use the Bezier spline with the control points as the concatenation of $\overrightarrow{0}$ with the other control points: $[\overrightarrow{0}, p_{1\cdots O-1}]$. Alternatively, we can compute $p_{0\cdots O-1}$ and use the Bezier spline with control points but subtract the first predicted point from all of them: $[p_{0\cdots O-1}]-p_0$, and the final change in position is $\Delta_0(x) = B_O(f(p)-f(p)_0,t)+p_0$. While both formulations are theoretically equivalent, we find it better to explicitly compute $p_0$, otherwise the initial frame will have deformations. In our evaluation, we do not subtract the first point at all, allowing movement in the first frame, but this can be subtracted out later from all points in a post-processing step.

\noindent
A diagram of the spline component for ray-bending can be seen in Fig.~\ref{fig:arch_diagram}.

Following NR-NeRF, we also learn how rigid each point in space is, allowing for efficient categorization of fixed regions. This rigidity is computed as a function of position:
\begin{align}
  \text{r}\in[0,1] =\sigma(\text{MLP}(x))\label{eq:rigidity_defn}
\end{align}
where $\sigma$ is defined as the sigmoid function $\frac{1}{1+e^{-x}}$, and this MLP is shared\footnote{This differs from NR-NeRF which uses two separate MLPs.} with computing the Bezier control points. Rigidity $\in[0,1]$ rescales the
difficulty of learning movement, making it easy to handle static scene
objects, where even slight motion would look incorrect in new views. The final change in position is defined as $\Delta(x) = \text{r}
\Delta_0(x)$.

In order to reconstruct RGB values, we also diverge from the original NeRF and NR-NeRF. Instead of only allowing for fully positional or view-dependent colors, we allow a small amount of linear scaling as a function of the view direction.
\begin{align}
  \text{RGB}_{pos} &= \sigma(\text{MLP}(x)) \nonumber \\
  \text{RGB} &= (1-\sigma(\text{MLP}(v))/2)\text{RGB}_{pos} \label{eq:refl_defn}
\end{align}
Because of the low number of
samples for a moving object at a given view, it is more difficult to learn specular
reflection, but it is often the case there are lighting changes which are necessary to model. Motivation for this modification can be found under the section on limitations(\ref{sec:refl_disc}).

\subsection*{Training}

For training, we sample random crops of random frames, computing the loss and back-propagating through both the NeRF and spline network. We use gradient descent to optimize control points and the canonical NeRF jointly, but note that there are also classical approaches to optimizing control points which could lead to faster optimization in the future. We use the Adam optimizer~\cite{Kingma2015AdamAM} with cosine simulated annealing~\cite{loshchilov2017sgdr} to go from $\num{2e-4}$ to $\num{5e-5}$ over the course of two days, and start with a low resolution $32\times32$ training image size as initialization before scaling to $256\times256$. We develop our approach on an NVIDIA GTX 1060, but run each experiment on one Tesla P100.

For some scenes, we are able to have higher learning rates at $\num{3e-4}$, but for much darker scenes it's necessary to lower the learning rate to $\num{1e-4}$ to converge, and find that if the scene is too dark, specifically the \textit{Hellwarrior} scene, we revert back to using only positionally dependent RGB, but still have difficulty converging since it is too dark.

Despite the guarantees of our method, it is still crucial to apply offset and divergence regularization defined in NR-NeRF~\cite{tretschk2021nonrigid} as:
\begin{align}
\ell_{\text{off}} = \frac{1}{|C|} \sum_{j\in C} \alpha_j \
    (\lVert \Delta_0(x_j) \rVert_2^{2-\text{r}}+\lambda_r \text{r})
\end{align}
\begin{align}
    \ell_{\text{div}} = \frac{1}{|C|} \sum_{j\in C} \omega_j' |\nabla\cdot(\Delta(x_j))|^2
\end{align}

Where $r$ is rigidity as defined in Eq.\ref{eq:rigidity_defn}, and $\lambda_r$ is a hyper-parameter, set to 0.3, and $\alpha_j$ refers to the accumulated visibility weight along a ray $T$, as defined in Eq.~\ref{eq:nerf}. We defer to NR-NeRF~\cite{tretschk2021nonrigid} for a complete explanation of these losses.

Our complete loss function is thus:

\begin{align}
    \ell = \sum\limits_{r\in\text{GT}}\lVert\text{GT} - f(r_o, r_d, t)\rVert_2^2 + \lambda_{\text{div}}\ell_\text{div} + \lambda_\text{off}\ell_\text{off}
\end{align}

Where we assign $\lambda_\text{div} = 0.3, \lambda_\text{off} = 30$, and $f$ is the rendering the described model at time $t$ with the rays from the known camera.

\subsection*{Voxel Spline-NeRF}

In addition to a model that uses an MLP to predict the control points, we demonstrate that using control points is also possible with a voxel-based approach, leading to much faster reconstruction times. Our formulation is identical to the MLP model, but instead of querying an MLP, the model trilinearly interpolates between the surrounding set of control points. We demonstrate the possibility of using a voxelized approach for reconstructing dynamic scenes in our experiments, but do not precisely measure how much faster it is than the MLP based approach, since it is heavily implementation-dependent, for example a voxel based approach would do well from using a handwritten CUDA extension, while our implementation is only written in Pytorch. We do note that training is faster and much less memory-intensive than the MLP based approach, allowing for an order of magnitude higher batch size while training, and converging faster. As compared to the MLP-based approach though, there is a degradation in quality. We expect there to be a need for additional regularization terms as compared to both previous voxel and dynamic reconstruction approaches.

Our voxel approach closely resembles the MLP based approach, only differing in storing a set of spline control points at every voxel position, as well as spherical harmonic coefficients in order to compute the linear color rescaling. Our simple implementation also does not differ significantly from our dynamic NeRF approach, differing by around 50 LOC. We defer to the supplementary material for results on the voxel model.

In order for our voxel approach to converge, we use losses introduced in NR-NeRF~\cite{tretschk2021nonrigid} and also find it necessary to apply total variational loss as used in Plenoxels~\cite{yu2021plenoxels}:

\begin{align}
    \ell_{TV} = \frac{1}{|V|} \sum_{v\in V} \sqrt{\sum_{d\in \{x,y,z\}} \Delta^2_d(v)}
\end{align}

Where $\Delta^2_d(v)$ is the difference between one of a voxel's value and one of its neighbor on the $d$ axis's corresponding values. This guarantees that there is relative consistency in the voxel grid. We note that we apply this to all components stored in the voxel grid, including the spline control points, rigidity, density, and the RGB. As in Plenoxels~\cite{yu2021plenoxels}, we stochastically sample this at each step.
\section*{Results}

In order to demonstrate our method, we run it on D-NeRF's~\cite{pumarola2020dnerf} synthetic dataset which contains 8 different rendered scenes with simple movement. These scenes have ground truth camera positions, viewing directions and timestamps. They capture physically plausible movement, without large discontinuities or jumps between frames.

We also demonstrate our method on a closed-room dataset, the Gibson dataset rendered for NeRFlow~\cite{du2021nerflow} using the iGibson environment~\cite{xu2019DISN}. This contains a single moving TurtleRobot from many similar views, similar to LLFF datasets.

We also note that we compare our method to our own implementation of NR-NeRF, to isolate the difference between our approach and using just an MLP. We make some modifications, by passing time explicitly rather than a latent vector, not varying regularization over epochs, and not requiring that at time $t=0$ we have no deformation in the rays.

\subsection*{Qualitative Results}

\begin{figure*}
    \includegraphics[width=\textwidth]{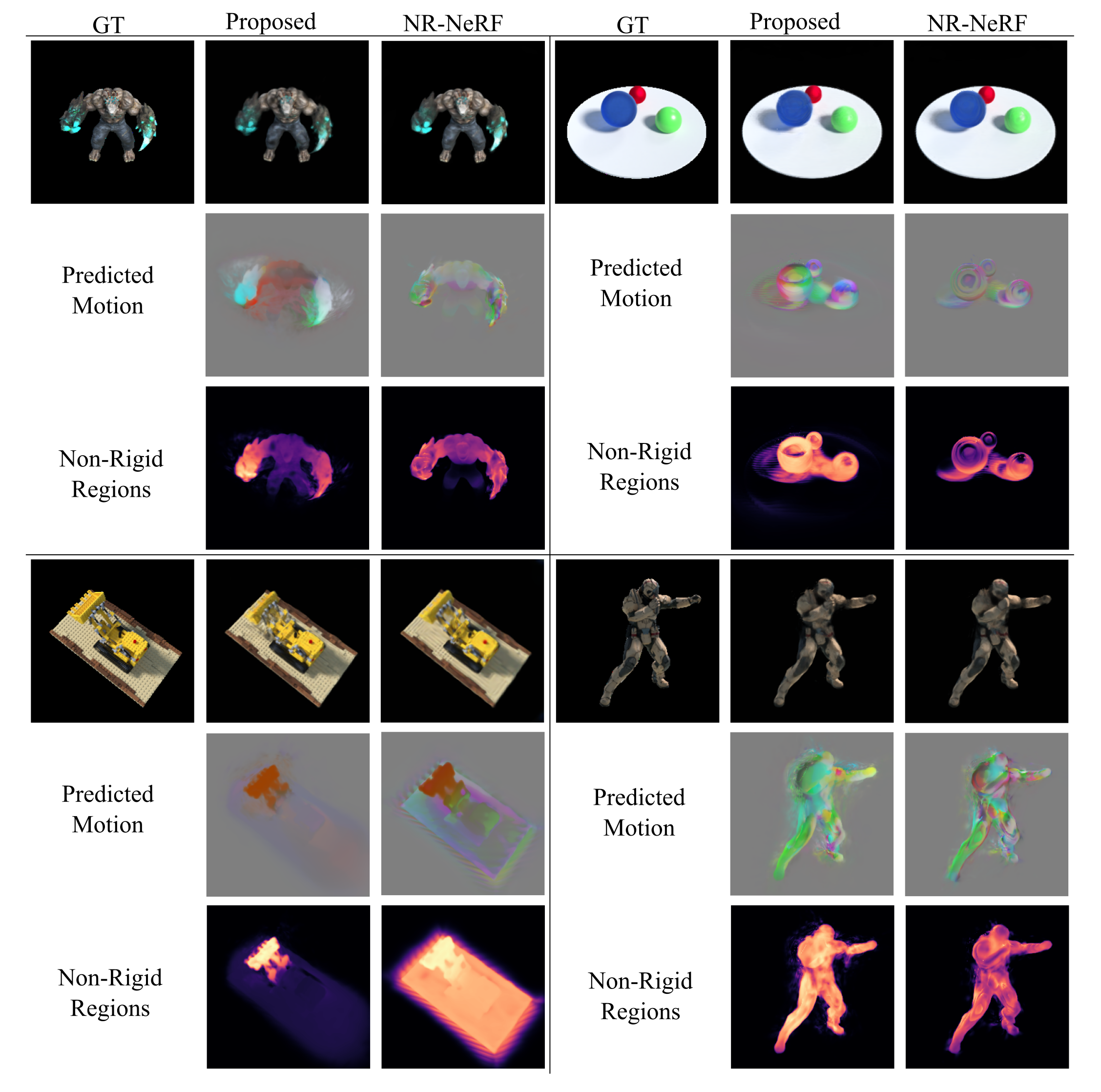}
    \caption{
        \label{fig:qual_cmp}
        \textbf{Visual comparison of Spline-NeRF to NR-NeRF.}
        Results comparing movement in our implementation of NR-NeRF~\cite{tretschk2021nonrigid} versus the proposed approach for using Bezier splines for modelling movement. The direction of motion is shown by color, distance is shown by intensity, and gray is 0 motion. Non-rigid regions are visualized as brighter. There is substantial difference in the predicted motion, and this can be seen in the Mutant and Hook scenes (top left, bottom right respectively), where NR-NeRF predicts very noisy 3D flow, but our approach is coherent, where coherence is loosely defined as being similar to nearby points. There is also a large difference in rigidity, where the spline has highly coherent non-rigid regions. This is especially noticeable in the Lego scene (bottom left), where NR-NeRF makes the entire model non-rigid, but adding a spline isolates the loader on the tractor to be non-rigid. We expect this is because splines are forced to have similar scale for movement, and the rigidity is less needed to compensate for error or differences in scale. For an MLP, it may predict different scales of motion for nearby points, but rigidity can compensate for this since it doesn't strictly learn 0 or 1.
        \
        The quality of the output is similar. In some portions, NR-NeRF captures higher quality output, such as in the shadows of the balls, or in the crispness of the Mutant's blue claws. On the other hand, the spline more accurately captures the knobs and shadows of the knobs on the Lego scene.
    }
\end{figure*}

The difference between our work and NR-NeRF can be observed in the difference of flow between scenes. It can be observed from Fig.~\ref{fig:qual_cmp} that our method captures coherent movement for objects, whereas for NR-NeRF movement may not be in the same direction, and we define coherence loosely as having similar movement within nearby space. For example, on the ball (top right), a significant portion does not appear to be moving. In addition, for the Lego scene (bottom left), our method isolates the loader on the tractor, whereas NR-NeRF cannot.

The difference between the two is also more clearly seen in videos of reconstruction. Spline-NeRF visibly has the effect of ``tweening'' between views, slowing into stops, while NR-NeRF appears less smooth.

\subsection*{Quantitative Results}

The qualitative comparison of our method to NR-NeRF is shown in Tab.~\ref{tab:dnerf_cmp}. Spline-NeRF is able to perform on par or with minimal degraded performance with our implementation of NR-NeRF on D-NeRF's synthetic dataset. This is likely because NR-NeRF does not impose constraints on the velocity or acceleration of movement, whereas Spline-NeRF is forced to create a smooth interpolation, which is more difficult. To be more precise, Spline-NeRF \textit{must} learn a continuous function, which is strictly more constrained than the set of functions that an MLP can learn, as the MLP can reproduce the observed views at each time, and implicit smooth between views. In practice NR-NeRF learns fairly smooth movement, but quantitatively looks different from our methods' movement, due to differences in velocity and acceleration. Bezier splines enforce that movement is fluid and can better reproduce in-between frames, trading off reproduction quality for smoothness. We expect that in longer sequences and data with larger gaps Spline-NeRF would benefit from this constraint.

\subsection*{Gibson Dataset}

\begin{figure*}[!ht]
    \includegraphics[width=\textwidth]{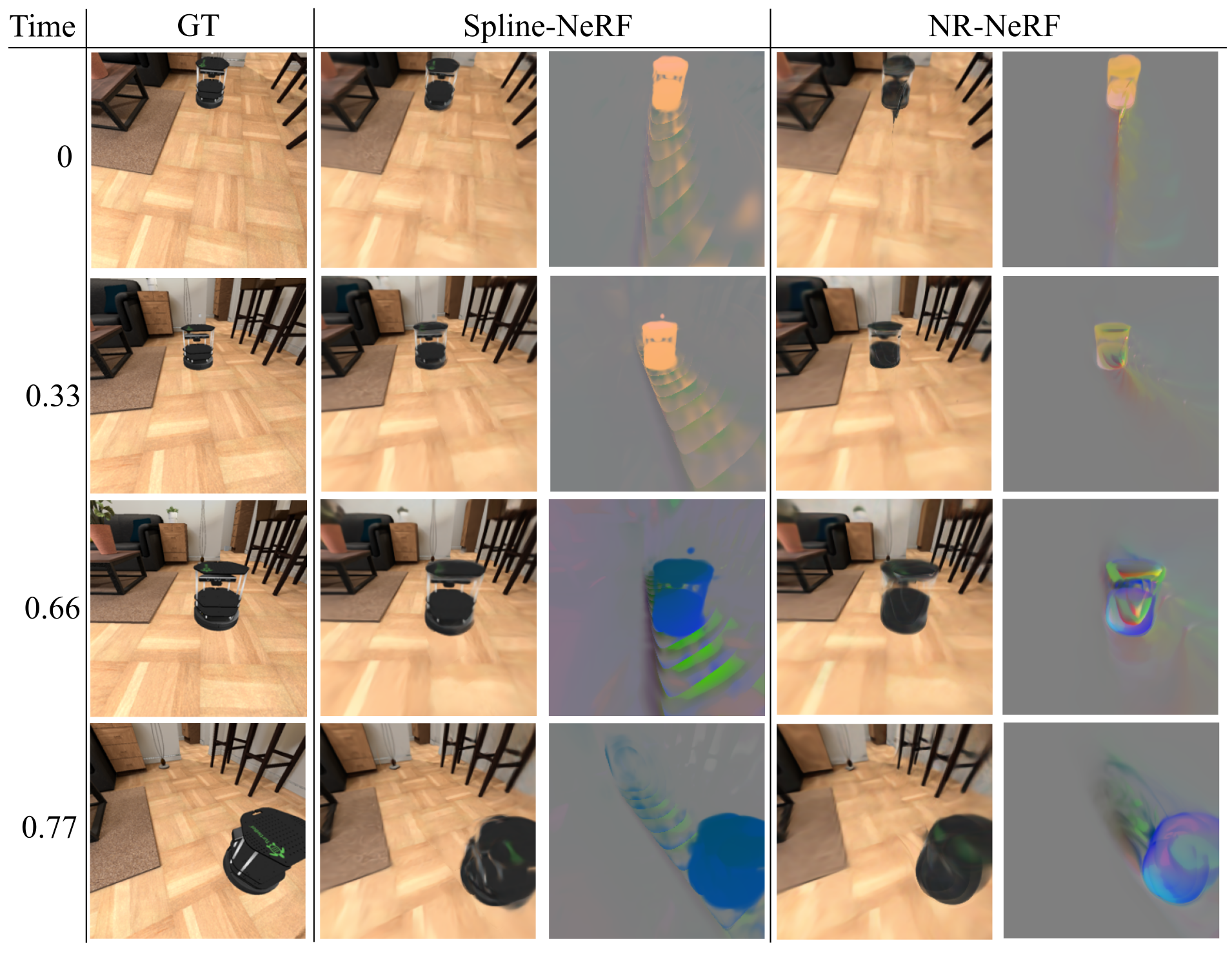}
    \caption{
        \label{fig:gib_cmp}
        \textbf{Visual comparison of Spline-NeRF to NR-NeRF on Gibson Dataset.} Our method produces coherent movement as compared to NR-NeRF, and is also visually more coherent, despite having a lower PSNR and MS-SSIM. It can be easily seen that the TurtleBot in NR-NeRF has artifacts, and does not move as a whole unit, whereas Spline-NeRF moves the whole robot uniformly. We do not include samples around $t=1$, since both our approach and NR-NeRF produce low-quality reconstructions around those times.
    }
    \vspace{-6mm}
\end{figure*}

We also include a more realistic dataset from NeRFlow~\cite{du2021nerflow}, the Gibson dataset, rendered from the iGibson environment~\cite{xu2019DISN}. Our method has median PSNR $24.537$ dB and $0.886$ SSIM, and NR-NeRF has median PSNR $24.591$ dB and $0.887$ SSIM. Instead of the mean, we use the median since there are test frames which contain an object close to the camera which is rarely seen in the training set, thus there are some frames with extremely low quality on both methods. Fig.~\ref{fig:gib_cmp} highlights the differences, notably, our method produces coherent movement, despite having lower quantitative metrics, and this can be seen in NR-NeRF's artifacts and the motion flow fields.

\begin{table*}[t]
    \centering
    \begin{tabular}{|c| c|c | c|c | c|c | c|c |}
    \hline
    \textbf{PSNR$^\uparrow$ $|$ MS-SSIM$^\uparrow$} & \multicolumn{2}{c|}{Bouncing Balls} & \multicolumn{2}{c|}{Hellwarrior$^\dagger$} & \multicolumn{2}{c|}{Hook} & \multicolumn{2}{c|}{Jumping Jacks} \\
    \hline
    NR-NeRF & \textbf{27.573} & \textbf{0.984}
           & 33.314 & 0.968
           & 27.954 & 0.978
           & 28.476 & 0.985 \\
    \hline
    Spline-NeRF & 26.418 & 0.979
               & 33.504 & 0.968
               & 28.104 & 0.979
               & 28.424 & 0.986 \\
    \hline
    & \multicolumn{2}{c|}{Lego} & \multicolumn{2}{c|}{Mutant} & \multicolumn{2}{c|}{Standup} & \multicolumn{2}{c|}{T-Rex} \\
    \hline
    NR-NeRF & 23.663 & 0.946
           & 30.382 & 0.989
           & 31.624 & 0.989
           & 26.649 & 0.985 \\
    \hline
    Spline-NeRF & 23.656 & 0.943
               & 31.183 & 0.992
               & 31.349 & 0.990
               & 26.056 & 0.982 \\
    \hline
    \end{tabular}
    \vspace{2pt}
    \caption{
        \label{tab:dnerf_cmp}
        \textbf{Comparison of mean PSNR and MS-SSIM for Spline-NeRF and NR-NeRF.} Bolded values are those that are significantly greater than the other.
        Bezier splines are able to recover movement with near equal accuracy in dynamic scenes as compared to NR-NeRF~\cite{tretschk2021nonrigid}. Due to the forced prior of continuous movement, we learn a smooth interpolation through each frame. We randomly samples all frames from the start of training. Here, we parametrize Spline-NeRF with 5 control points. \newline
        $^\dagger$We had difficulty consistently reproducing results on this dataset. This is mostly because it is extremely dark: it is difficult to distinguish the black background from the object.
    }
    \vspace{-6mm}
\end{table*}

\section*{Discussion}

Learned Bezier splines produce coherent movement in objects and are able to converge with equal to speed as prior work. Our method demonstrates smooth interpolation, as enforced by the structure of our approach. We are thus able to super-sample between frames at arbitrary resolution while guaranteeing smoothness. In addition, because of the analytic nature of splines, we are better able to understand the movement of objects within a scene.

Another computational benefit of our approach is that for a single view, we only need to evaluate the deformation MLP \textit{once} to compute the Bezier spline at each point, then we can sample the Bezier spline at an arbitrary number of times to compute a video. This should allow for efficient single view video reconstruction, but we leave future work for this.

Finally, even without the above optimization, our model runs approximately as fast as NR-NeRF, but can be optimized further, either by introducing voxels, or using other approaches such as a different optimization scheme for splines.
\section*{Limitations}

\paragraph{Reduced PSNR.} Our work reduces the PSNR and MS-SSIM on the test set slightly. This is due to the requirement for a continuous function, which is more difficult to learn than directly predicting from an MLP, which from our qualitative results may not produce coherent movement or isolate non-rigid regions. This can be considered a trade-off between bias and variance, where we reduce the space of learned functions, in order to enforce that they are physically plausible. We argue that this trade-off may be beneficial in some cases, for example if we are interested in controlling movement, it's easier to modify our method's control points than it is to modify the output of the MLP. Our method is also easily interpretable and provably correct and for certain kinds of applications, these qualities would be preferable to quality. For applications which purely focus on reconstruction quality, and do not care about coherent movement or interpretability, it may make more sense to use an MLP.

\paragraph{Bezier Hyperparameters.} While our work is able to capture 3D movement, there are still limitations in what movement it can capture. Specifically, the number spline points determines the degrees of freedom of movement, and correctly selecting the number can be challenging. If too many points are selected, it can suffer from oscillations, akin to Runge's phenomenon, and this was observed during experimentation. If too few points are selected, the space of learnable movement is significantly smaller. For the datasets used and those examined in prior work, simple movement is learned, which protects against these issues by selecting between 4-6 control points, but for more complex movement it is be more difficult.

\paragraph{Transient Content.} This work also does not tackle transient content in a scene. For example, effects such as fire or complex lighting changes beyond moving shadow cannot be accurately captured. This is not a focus of this work, but is orthogonal and important when broadly considering dynamic scenes. While this work does not handle this, in contrast to prior work which may enable learning this by teleporting particles with the deformation network, our approach \textit{prevents} modelling those effects, which is beneficial since an orthogonal component may be able to learn them more effectively.

\paragraph{Reflectance}\label{sec:refl_disc} Another change in our approach is that the reflectance model defined in Eq.~\ref{eq:refl_defn} may not be clearly motivated. It is able to model more than prior work, by limited linear scaling. This may be seen as a pro and con, in that we may be overfitting to the specific dataset we are testing on, but we argue it is broadly viable. In fact, the necessity for this reflectance is that there are non-negligible view-dependent effects in the dataset which a purely positional model cannot capture, but a fully view-dependent model cannot generalize to. Ours fits in the middle, capturing just what is necessary, while preventing content from becoming completely dark from other views. The idea behind this approach was learning a general gamma correction: $A (\text{RGB})^\gamma$, but found that learning $\gamma$ even in the limited range $[0.5,1.5]$ failed due to instability. Thus, we only retained the linear term, and found this to improve performance. We can think of this as a diffuse BSDF, which has an explicit term for diffusion $V\cdot L$, where in our case L is unknown but fixed. In addition, our limitation of $A\in[0.5,1]$ can be considered an implicit bias on the existence of global illumination, in that no particle can be fully black. Hopefully, this sufficiently motivates the modification to prior work's RGB prediction.

\section*{Future Work}

One large next step in neural rendering is to encode dynamic models on top of highly-efficient NeRF representations, such as using a Plenoxel~\cite{yu2021plenoxels}, Instant Neural Graphics Primitives~\cite{mueller2022instant} or other structure to allow for rapid reconstruction of dynamic scenes. Our voxel implementation does not utilize sparsity as much as possible, nor does it use modifications such as hash-encoding, partially to demonstrate the efficacy of our approach on its own two feet. To the author's knowledge, there is no real-time construction of 3D scenes since there did not exist a classical approach to reliably reconstruct movement. Bezier splines may help fill in this gap, requiring a small number of parameters and allowing for efficient rendering and training without requiring a costly MLP evaluation. If this follows the trend of scene reconstruction, it may be multiple orders of magnitude faster to reconstruct dynamic scenes, without loss in quality, and we hope that this work gets adopted for this purpose.

In addition, long duration (minutes or hours) dynamic scene reconstruction has not yet shown to be plausible, but Bezier splines are easily extendable to long scenes by turning them into poly-Bezier splines or adding more control points. It is not immediately clear how to enable efficient reconstruction over long sequences, but using splines is a clear avenue for future work for learning long dynamic scenes with continuity guarantees.

We also hope that future work explores variations on spline formulations, as we select
Bezier splines due to simplicity, numerical stability and efficiency of evaluation. It may be that other splines might have stronger expressivity for reconstruction or other desirable properties, and thus may be useful in different contexts.

\section*{Conclusion}

In conclusion, we devise a new architecture for $C^0, C^1$ continuous interpolation, and show that it works with dynamic NeRF, performing on par with prior work. Our architecture is able to accurately reconstruct scenes, while providing strong smoothness guarantees using a well-studied tool. This leads to coherent movement, which can be observed in a reconstructed video. Hopefully, this inspires more use of classical tools inside of the differentiable rendering pipeline, so we can accurately and efficiently recover physical phenomenon.

While our work is incremental, requiring very little modification to existing code, since it changes the underlying structure to an analytic form, so we expect a large amount of tooling and analytical tools can be built on top of this change, leading to better understanding and analysis of dynamic 3D content.

{\small
    \bibliographystyle{splncs04}
    \bibliography{ref}
}

\end{document}


\pagestyle{headings}
\mainmatter
\def\ECCVSubNumber{6195}

\title{Spline-NeRF: Supplement}
\author{Anonymous ECCV Submission}
\institute{Paper ID \ECCVSubNumber}

\maketitle

\section*{Additional Qualitative Results}

\begin{figure*}
    \centering
    \includegraphics[width=\textwidth]{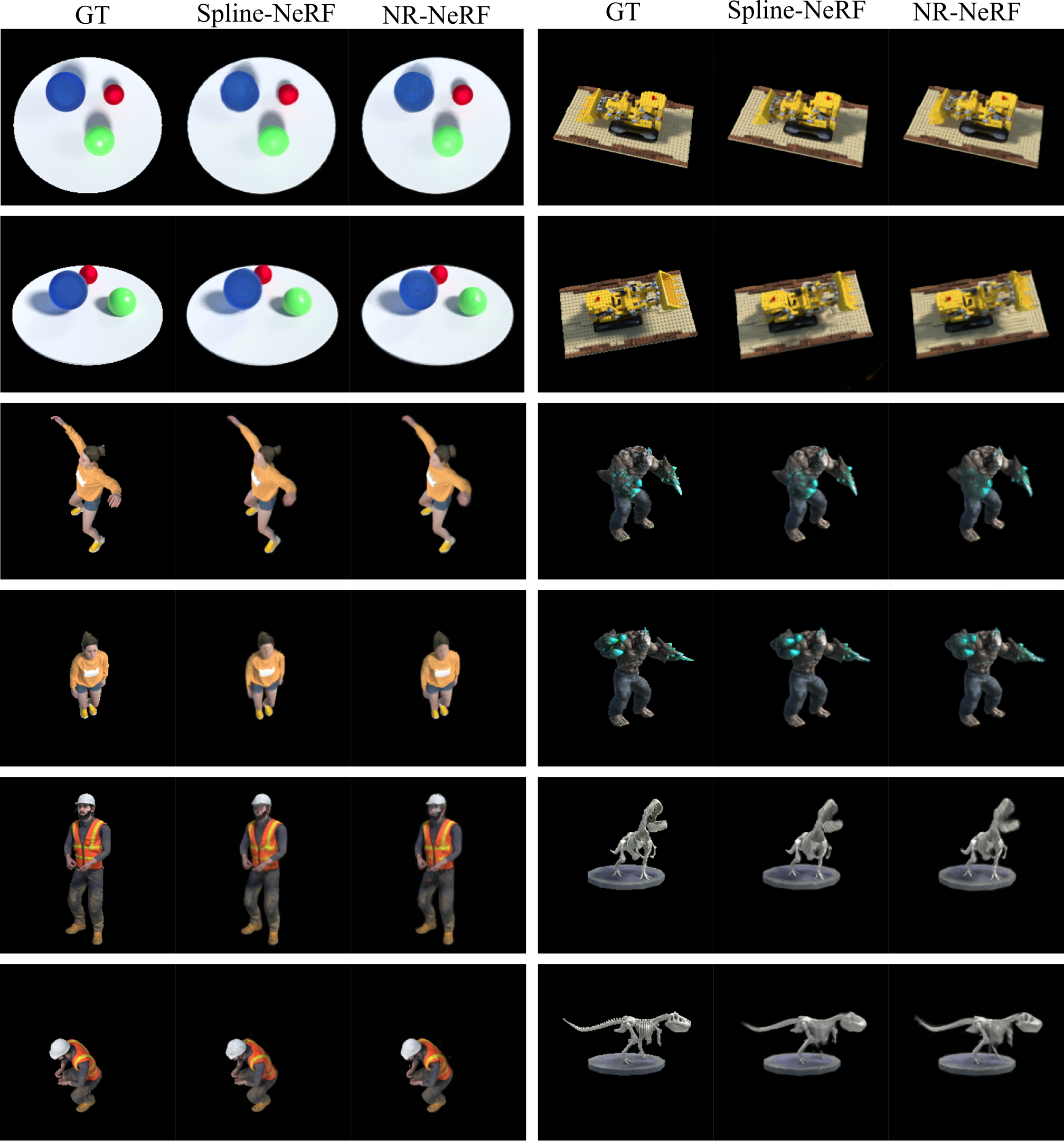}
    \caption{
        \label{fig:dnerf_more}
        \textbf{Additional test frames from D-NeRF dataset.} More qualitative comparisons of our method to NR-NeRF. There are some cases where NR-NeRF can capture more high-quality details than Spline-NeRF, but also cases where Spline-NeRF is able to more accurately reconstruct the output. These manifest in high-frequency details, but the low-frequency components of both reconstructions are approximately the same.
    }
\end{figure*}

We include additional qualitative results in Fig.~\ref{fig:dnerf_more} to allow for comparison of the two methods. We include these additional results in order to better visualize the qualitative differences in our method, which cannot be seen through quantitative results alone.

\section*{Voxel Spline Results}

\begin{figure}
    \vspace{-6mm}
    \centering
    \begin{minipage}[c]{0.5\textwidth}
    \includegraphics[width=\textwidth]{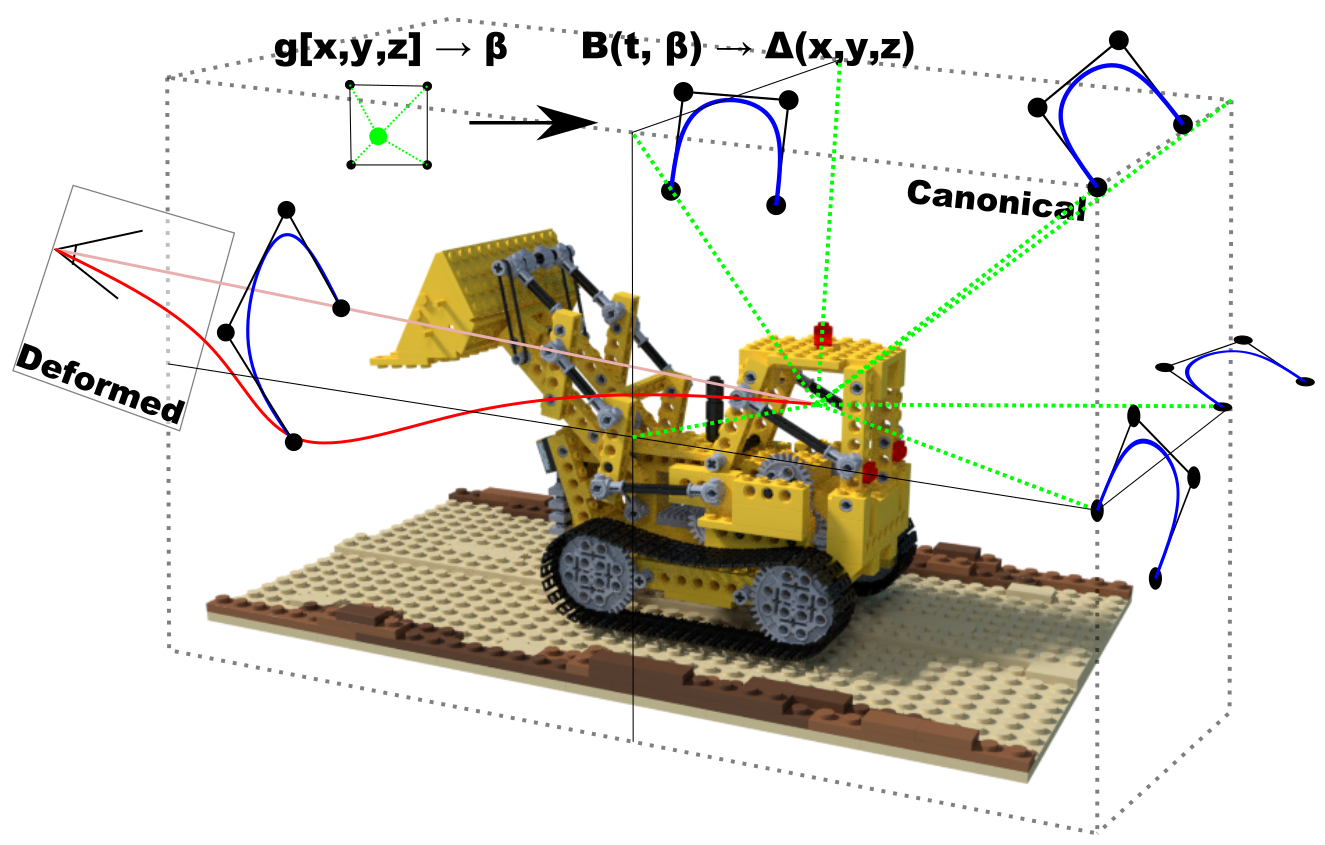}
    \end{minipage}
    \begin{minipage}[c]{0.45\textwidth}
    \caption{
        \label{fig:voxel_diagram}
        \textbf{Voxel implementation of Spline-NeRF.}
        Our implementation of Spline-NeRF does not deviate much from our MLP-based implementation, but instead stores control points at each voxel coordinate.
    }
    \end{minipage}
    \vspace{-10mm}
\end{figure}

\begin{figure*}
    \centering
    \includegraphics[width=\textwidth]{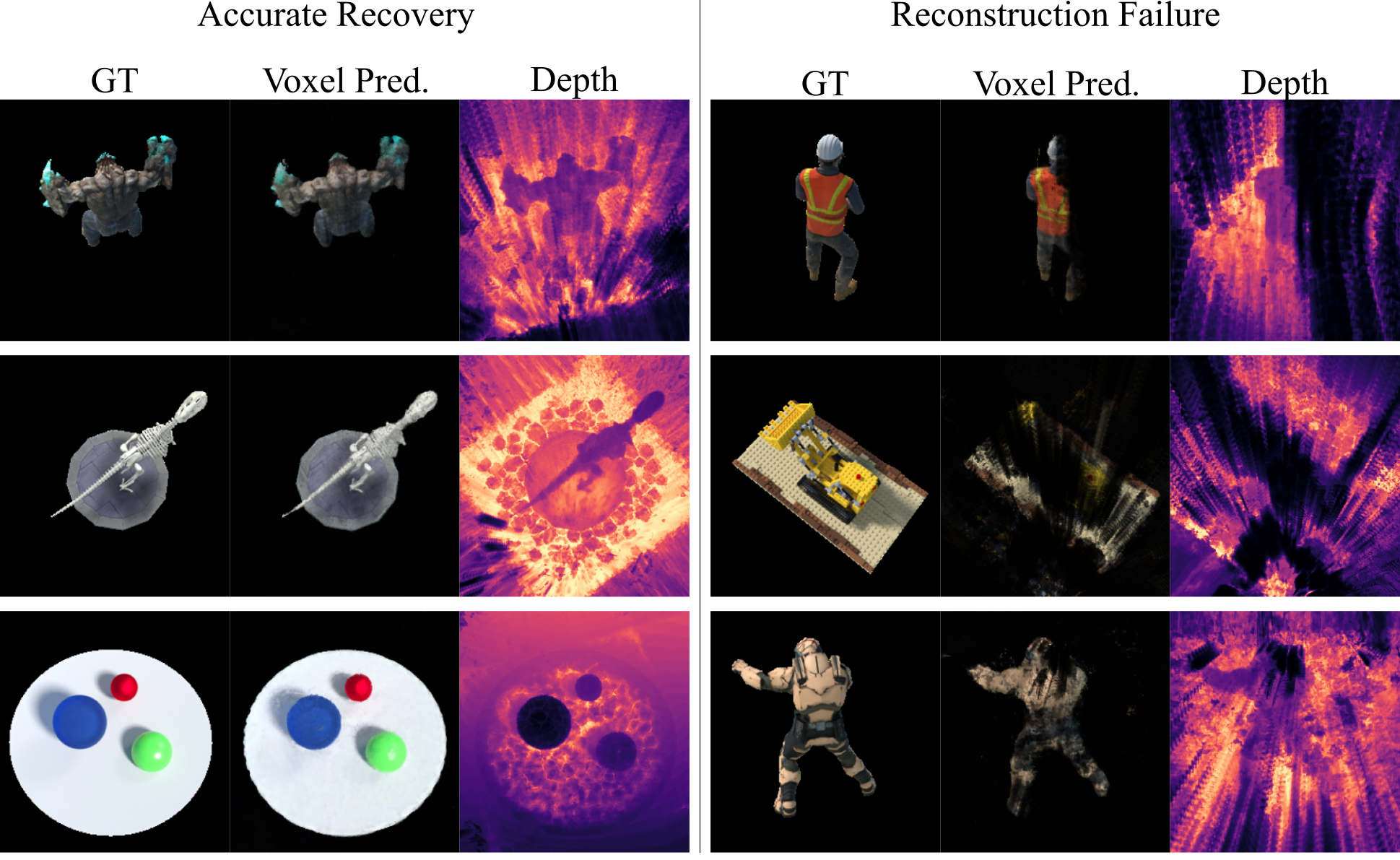}
    \caption{
        \label{fig:voxel_results}
        \textbf{Results from Voxel Spline-NeRF.} Results of reconstruction using a Voxel implementation of Spline-NeRF. We include both successful output from the voxel implementation, which is cherry-picked to showcase the potential of the method, as well as failure cases in order to show that there are still shortcomings without additional regularization. We note that even in the successful cases, it's clear from the visualized depth that there are artifacts, but they are black and thus do not affect the output.
    }
\end{figure*}

We also include some results from our naive voxel implementation in Fig.~\ref{fig:voxel_results}. We pick a few successful results to highlight that a voxel implementation can accurately recover a moving scene, but also show some failure cases to highlight that a naive voxel implementation may need additional regularization. Notably, there are a large amount of floating black spots, which is not necessarily a problem in a static reconstruction if they are not observed, but may be more of a problem if they are \textit{moved} into the line of sight. Thus, we suspect that regularization on unobserved voxels may be necessary.

Since a complete voxel implementation is not the focus of this work, we leave this for future work, but highlight its potential for fast reconstruction of dynamic scenes.


